# A Study on the MCP × A2A Framework for Enhancing Interoperability of LLM-based Autonomous Agents

Dr. Cheonsu Jeong

Hyper Automation Team, SAMSUNG SDS, Seoul, 05510, South Korea

## ABSTRACT

This paper provides an in-depth technical analysis and implementation methodology of the open-source Agent-to-Agent (A2A) protocol developed by Google and the Model Context Protocol (MCP) introduced by Anthropic. While the evolution of LLM-based autonomous agents is rapidly accelerating, efficient interactions among these agents and their integration with external systems remain significant challenges. In modern AI systems, collaboration between autonomous agents and integration with external tools have become essential elements for building practical AI applications. A2A offers a standardized communication method that enables agents developed in heterogeneous environments to collaborate effectively, while MCP provides a structured I/O framework for agents to connect with external tools and resources. Prior studies have focused primarily on the features and applications of either A2A or MCP individually. In contrast, this study takes an integrated approach, exploring how the two protocols can complement each other to address interoperability issues and facilitate efficient collaboration within complex agent ecosystems.

## I. INTRODUCTION

RECENT advancements in artificial intelligence (AI) have highlighted the importance of large language models (LLMs) across various domains [1]. Generative AI, which can create new content such as text, images, audio, and video based on extensive training data, enables users to leverage AI services with ease [1], [2]. In particular, the development of AI has led to a growing utilization of autonomous agents—AI systems capable of independently performing tasks, solving problems, and supporting decision-making based on user inputs. However, complex real-world problems often exceed the capabilities of a single agent. This has given rise to the necessity of multi-agent systems (MAS), where multiple agents collaborate to overcome such limitations. With the emergence of LLMs, LLM-based MASs have gained attention as a new paradigm [4]. Unlike traditional MASs, these systems enable more flexible and adaptive collaboration through natural language processing. Each agent can specialize in a specific domain while coordinating complex tasks through natural language interaction [5].

For agents to be effectively utilized in real-world environments, seamless integration with external tools, APIs, and databases is critical. To address this, Anthropic released the Model Context Protocol (MCP) as open source in November 2024 [6]. Similarly, efficient communication and collaboration among agents is a prerequisite for successful MAS deployment. However, agent systems developed by different frameworks and vendors have faced interoperability limitations. To resolve this, Google released the Agent-to-Agent (A2A) protocol as an open source project in April 2025 [7]. A2A provides a standardized communication framework for agents to collaborate across different environments, while MCP offers a structured I/O system that enables agents to interact with external tools and resources [8]. Although these protocols function independently, they serve complementary roles and form a foundational basis for building practical agent ecosystems.

This paper analyzes the technical architecture and operational principles of A2A and MCP, exploring how they synergistically contribute to constructing scalable and interoperable agent ecosystems. By presenting an integrated architecture using LangGraph and detailing key implementation logic, we demonstrate how theoretical concepts are translated into real-world applications. Through this combined use of A2A and MCP, we propose a practical approach to enhancing interoperability and development efficiency of LLM-based agent systems across various enterprise environments, thereby unlocking the potential to transition agent technologies from experimental stages to fully deployed services.

## II. RELATED WORK

### A. Concept and Evolution of Autonomous Agents

An automation agent refers to more than just a task executor—it is an intelligent software robot that integrates AI technologies to perceive complex situations, analyze data, interact with users, and even improve its performance through learning [9]. Autonomous agents are systems that perceive their environment, autonomously set goals or formulate plans to achieve predefined objectives, and make decisions and act independently by learning and adapting [10], [11]. These agents can respond flexibly to unpredictable or dynamic environments, solving problems on their own as needed. The recent advancement of large language models (LLMs) has significantly expanded the capabilities and application areas of autonomous agents. They are now widely used in information retrieval, data analysis, decision support, and task automation, and are increasingly being adopted as productivity tools in enterprise environments.

Autonomous agents possess several key characteristics, as summarized in Table 1:

\* Corresponding authors:
E-mail address: csu.jeong@samsung.com (C. Jeong).

TABLE I. KEY CHARACTERISTICS OF AUTONOMOUS AGENTS

| Item | Description |
|---|---|
| Autonomy | The ability to operate independently without external intervention. |
| Goal-Oriented | The ability to formulate and execute plans to achieve specific objectives. |
| Adaptability | The capability to adjust behavior in response to environmental changes. |
| Social Ability | The ability to interact with other agents or users. |

However, due to constraints in knowledge, functionality, and resources, a single agent often faces limitations in solving complex problems. To overcome these limitations, Multi-Agent Systems (MAS) have emerged, where multiple agents collaborate toward a shared goal.

## B. Necessity of Multi-Agent Systems (MAS)

A MAS is a framework in which multiple autonomous agents collaborate to solve complex problems. Each agent is assigned a specific role and area of expertise, and their cooperation enables the resolution of problems that would be infeasible for a single agent alone [1].

The main advantages of MAS include:

1. **Distributed Problem Solving:** Complex problems are decomposed into sub-problems, with each agent addressing a specific domain.
2. **Scalability:** System functionality can be expanded by adding or modifying agents.
3. **Robustness:** System functionality is maintained even if some agents fail.
4. **Diverse Perspectives:** Agents with different expertise analyze problems from varied angles.

However, effective MAS operation requires seamless communication and collaboration among agents, necessitating standardized communication protocols and collaboration mechanisms.

## C. Importance of Agent Communication Protocols

Agent communication protocols define how agents exchange messages, including the format, communication rules, and semantic interpretation. Communication is fundamental to MAS, enabling autonomous entities to coordinate and collaborate in solving complex problems [12].

An effective communication protocol must satisfy the following requirements:

1. **Interoperability:** Support communication between agents developed on different platforms and frameworks.
2. **Extensibility:** Accommodate new functions and requirements.
3. **Security:** Provide mechanisms for secure communication.
4. **Efficiency:** Minimize communication overhead and support efficient message exchange.
5. **Standardization:** Offer standardized formats and rules for broad adoption.

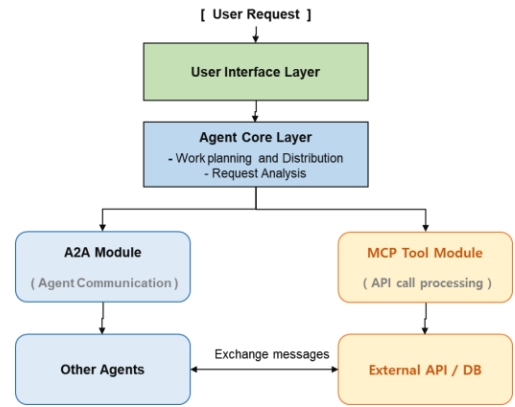

Fig. 1. Integrated Implementation Architecture Concept

## D. Technical Analysis of the A2A Protocol

### 1. Concept and Structure of A2A Protocol

Only two levels of headings should be numbered. Lower level headings remain unnumbered; they are formatted as run-in headings.

The A2A protocol is a standardized communication framework designed to enable effective collaboration among autonomous agents developed in diverse environments. Released as open-source by Google, it abstracts the complexity of agent communication and provides a consistent interface to ensure interoperability [8].

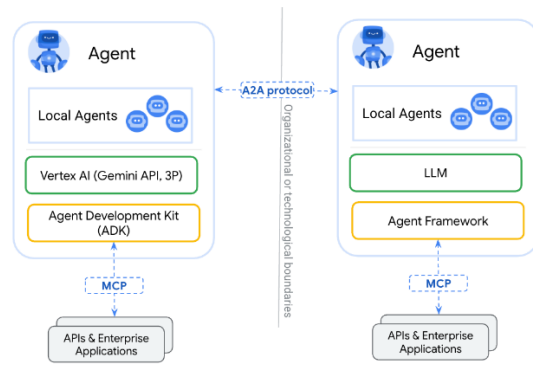

Fig. 2. Open Standard A2A Protocol Concept Diagram [7]

The core structure of the A2A protocol consists of:

1. **Agent Card:** Metadata in JSON format specifying an agent's functions, roles, and capabilities, enabling other agents to discover and determine appropriate collaboration methods.
2. **Message Format:** Defines a structured format for agent communication, including headers, body, and parts, supporting various content types (text, images, structured data).
3. **Task Management:** Adopts a task-based communication model, tracking task status, progress, and outcomes.
4. **Artifacts:** Standardized data structures representing task outcomes, such as documents, code, or images.

The A2A protocol achieves loose coupling and high cohesion, enhancing the flexibility and scalability of agent systems.

### 2. Core Functions of the A2A Protocol

A2A supports a range of core functionalities that enable collaborative operations among agents:

1. **Capability Discovery:** Agents advertise their capabilities using an "Agent Card" in JSON format. This mechanism functions similarly to service registries in Service-Oriented Architecture (SOA), allowing for dynamic discovery and coordination [8]. Each Agent Card includes essential information such as agent ID, name, description, supported functions, input/output formats, authentication requirements, and version information. This mechanism supports dynamic expansion of the agent ecosystem, enabling existing agents to recognize and collaborate with newly added agents automatically.
2. **Task and State Management:** A2A structures agent communication around tasks, each with a unique identifier and life cycle states (e.g., created, in-progress, completed, failed). Results are returned as standardized artifacts [8]. This task-oriented design supports asynchronous processing, real-time monitoring, task interruption and resumption, and consistent handling of results. It facilitates coordination in complex collaboration scenarios and provides a foundation for recovery strategies in case of failure.
3. **Secure Collaboration:** A2A includes mechanisms to ensure secure communication between agents. It supports enterprise-grade authentication and OpenAPI-based authorization, enabling safe exchanges of context, responses, artifacts, and user directives [8], [13], [14]. Key security elements include authentication, authorization, encryption, access control, and audit trails. These features allow for secure inter-agent communication while maintaining data protection, which is essential for enterprise use cases.
4. **User Experience Negotiation:** Each A2A message can include fully composed "parts" that represent complete content blocks—such as formatted text, images, or interactive elements. The protocol supports negotiation for user interface elements, enhancing user experience consistency across different clients [8]. This feature provides benefits such as multimodal content support, consistent UX delivery, adaptation to various client environments, and standardized representation of interactive content. It ensures consistent rendering of agent-generated content across interfaces.

### 3. Technical Advantages of the A2A Protocol

The A2A protocol provides the following technical advantages:

1. **Standardized Communication:** Standardizes communication among agents developed in different frameworks and vendors, ensuring interoperability [8], [13].
2. **Loose Coupling:** Agents can collaborate without knowledge of each other's internal implementation, enhancing system flexibility and scalability.
3. **Dynamic Discovery:** Dynamic capability discovery through agent cards enables easy integration of new agents and functionalities.
4. **Multimodal Communication:** Supports exchange of various types of content, including text, images, and structured data.
5. **Extensible Design:** JSON-based message format and modular structure allow easy accommodation of new requirements and functionalities.
6. **Enterprise Support:** Provides essential security features such as authentication, authorization, and auditing for enterprise environments.

These technical advantages significantly reduce the development and deployment costs of agent systems and improve collaboration efficiency among diverse agents.

### E. Technical Analysis of the Model Context Protocol (MCP)

### 1. Concept and Structure of MCP

The MCP is a standardized protocol designed to enable AI agents to connect effectively with external tools, APIs, and resources [14]. As a core component enhancing agent practicality, MCP establishes a structured input/output framework that facilitates seamless interactions between agents and diverse external resources [8].

The primary structural elements of MCP include:

1. **Context Definition:** Specifies the context in which the agent interacts with external tools, including tool functionalities, input parameters, and output formats.
2. **Schema-Based Interface:** Uses JSON Schema to clearly define the input and output types for each tool. This allows the agent to understand the tool's interface and provide appropriate input.
3. **Function Calling Mechanism:** Provides a standardized mechanism for agents to invoke external tool functions and retrieve their results.
4. **State Management:** Maintains state information during tool usage, supporting complex multi-step interactions.

The structure of MCP balances clearly defined boundaries between agents and tools with the flexibility necessary for dynamic interaction.

### 2. Core Functions of MCP

MCP provides various core functions for agents to interact effectively with external tools:

1. **Tool Discovery and Description:** MCP provides a mechanism to clearly describe tool functionality and usage. This is achieved through JSON schema-based tool descriptions, which include tool name and description, list of supported functions, input parameters and types, output formats and types, examples, and usage instructions [8]. This tool discovery mechanism enables agents to dynamically discover available tools and select appropriate tools to perform tasks.
2. **Function Calling and Result Handling:** MCP provides a standardized way for agents to call functions of external tools and handle results. This process consists of the following steps [8]:
    - Function Selection: The agent analyzes the user's request and selects an appropriate function.
    - Parameter Construction: The agent constructs the parameters required for function execution.
    - Function Call: The agent calls the function with the constructed parameters.
    - Result Reception: The agent receives the result of the function execution.
    - Result Interpretation: The agent interprets the received result and provides it to the user or uses it for subsequent tasks.

This function calling mechanism enables agents to utilize various external tools and APIs to perform actual tasks.

TABLE 2. COMPARISON OF MCP, A2A, AND GENERAL API

| Feature | MCP | A2A | General API |
| --- | --- | --- | --- |
| Primary Purpose | Connecting agents to external tools/services | Communication and collaboration between agents | Data exchange between client and server |
| Communication Direction | Agent → Tool | Agent ↔ Agent | Client ↔ Server |
| Message Format | JSON Schema-based | Structured messages (Header, Body, Parts) | Varies: REST, SOAP, etc. |
| State Management | Context-based state tracking | Task-based state management | Stateless or session-based |
| Discovery Mechanism | Tool Descriptions | Agent Cards | API documentation, OpenAPI specs |
| Authentication Method | Supports various mechanisms | Enterprise authentication, OpenAPI-based | Varies: API Key, OAuth, etc. |
| Error Handling | Standardized error codes and recovery strategies | Task failure handling and recovery | HTTP status codes, error responses |
| Scalability | Easy integration of new tools | Easy integration of new agents | Scalable through endpoint expansion |
| Implementation Complexity | Moderate | High | Low to moderate |
| Multimodal Support | Limited | Comprehensive (Text, Image, Structured Data) | Limited |
| Level of Standardization | Emerging standard | Emerging standard | Widely varied |
| Developer Experience | Self-descriptive, schema-based | Discovery via Agent Cards | Highly documentation-dependent |
| Real-time Interaction | Limited | Supported | Depends on API design |
| Security Model | Tool-specific access control | Inter-agent authentication and authorization | Endpoint-specific access control |
| Typical Use Cases | Tool/API connection, automation via function calls | Collaborative problem solving, long-running coordination | Information retrieval, command dispatch |

3. **Context Management:** MCP provides functions to effectively manage context during interaction between agents and tools. This includes session management, state tracking, context propagation, and memory management [8]. These context management functions enable agents to maintain consistent state and interact with tools during complex multi-step tasks.
4. **Error Handling and Recovery:** MCP provides mechanisms to handle and recover from errors that may occur during tool usage. This includes standardization of error codes and messages, retry strategies, fallback paths, and graceful degradation [8]. These error handling mechanisms enable agents to operate robustly even in unexpected situations.

*3. Technical Advantages of MCP*

MCP provides the following technical advantages:
1. **Standardized Tool Integration:** Enables consistent integration of various tools and APIs, making it easy to extend agent functionality.
2. **Type Safety:** JSON schema-based interfaces ensure type safety, reducing errors in agent-tool interaction.
3. **Self-documenting:** Tool descriptions are included as part of the protocol, allowing tool usage to be understood without separate documentation.
4. **Extensible Design:** Provides an extensible structure for easily adding new tools and functionalities.
5. **Language and Platform Independence:** Adopts an open standard that can be implemented in various programming languages and platforms.
6. **Developer-friendly Interface:** Provides an intuitive interface that is easy for developers to understand and implement.

These technical advantages significantly expand the practical utility and scope of agents and make it easier for developers to build and extend agent systems.

*F. Comparison of MCP, A2A, and API*

The main characteristics of MCP, A2A, and general APIs are compared in Table 2.

This comparison table shows the main characteristics and differences of each protocol/interface. MCP focuses on enabling agents to interact effectively with external tools, A2A facilitates collaboration among agents, and general APIs provide a standard interface for data exchange between services. These three approaches are complementary and, when used together, can build a powerful agent ecosystem.

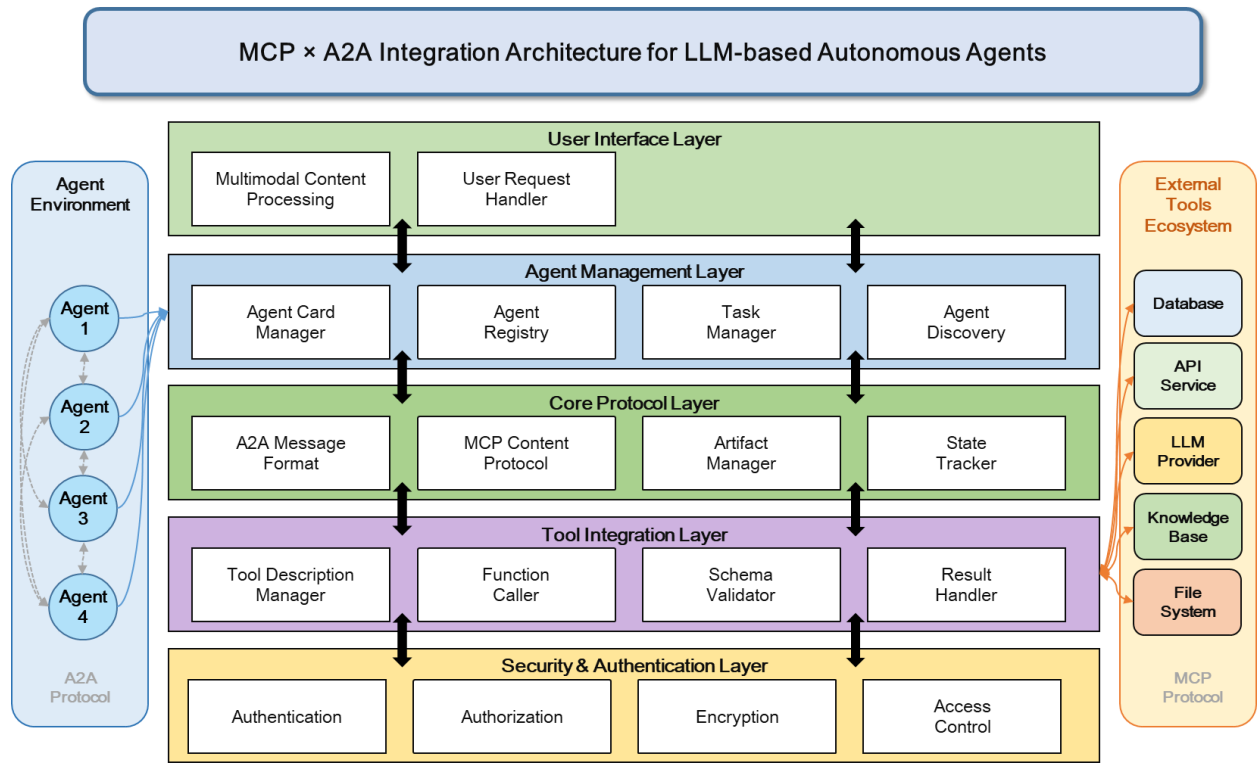

Fig. 3. The MCP×A2A Framework integrates agent-to-agent communication with tool connectivity in a layered architecture

## III. MCP–A2A INTEGRATED IMPLEMENTATION METHODOLOGY

### A. Implementation Architecture Overview

To implement MCP and A2A in real systems, a systematic architectural design is required. This section explains the architectural components and design principles for effectively implementing both protocols.

The integrated implementation architecture of MCP and A2A consists of the following main layers (Fig. 3):

1. **User Interface Layer:** Responsible for interaction between users and the agent system, handling user requests and displaying results. This layer provides a consistent user experience across diverse client environments and effectively represents multimodal content using A2A's user experience negotiation features.
2. **Agent Management Layer:** Manages the lifecycle, state, and functionality of agents, including creation and management of agent cards and coordination of agent communication. According to Habler et al. [13], this layer creates a flexible collaboration environment through dynamic agent discovery and capability negotiation.
3. **Core Protocol Layer:** Implements the basic protocol specifications of A2A and MCP, providing core functions such as message formats, communication rules, and error handling. This layer ensures protocol compliance and enables consistent implementation across diverse environments.
4. **Tool Integration Layer:** Integrates external tools and APIs through MCP, managing tool descriptions and processing function calls. According to the latest specification of the MCP [14], this layer provides type safety and self-documenting features through JSON schema-based interfaces.
5. **Security and Authentication Layer:** Responsible for securing agent communication and tool usage, handling authentication, authorization, and data encryption. According to Cloud Security Alliance [15], this layer builds a secure agent collaboration environment using OAuth 2.0-based authorization mechanisms and end-to-end encryption.

This layered architecture improves system maintainability and scalability through separation of concerns. Each layer can be developed and tested independently and can be extended or replaced as needed, allowing the architecture to evolve flexibly with the agent system.

### B. A2A Implementation Procedure

The implementation of the A2A protocol involves a structured sequence of steps, including the definition of agent metadata, the setup of communication logic, and task management. These steps ensure that agents can interact effectively in a standardized and interoperable manner.

#### 1. Agent Card Definition

An agent card is JSON-formatted metadata that specifies an agent's functionality and role. Agent card definition consists of the following steps [8]:

1. **Define Agent Identifier and Basic Information:** Define basic information such as agent ID, name, description, and version.
2. **Define Capabilities:** Define the list of functions provided by the agent, including parameters and return values for each function.
3. **Define Authentication Requirements:** Define authentication

methods and permission scopes required for agent usage.

4. **Add Metadata:** Define additional metadata such as agent category, tags, and creator information.

Agent cards, as JSON-formatted metadata, define agent ID, function list, and authentication requirements, enabling other agents to discover and utilize them [8]. According to Salimbeni [16], standardized agent cards support the dynamic scalability of the agent ecosystem, allowing existing agents to automatically recognize and collaborate with new agents added to the system.

### 2. Message Processing Implementation

A2A message processing is implemented through the following steps [8]:

1. **Message Reception:** Receive messages from other agents through various communication channels such as HTTP requests, message queues, or WebSocket.
2. **Message Parsing:** Parse the received message to extract headers, bodies, and parts. A2A messages are structured in JSON format, making parsing easy.
3. **Message Validation:** Validate the message format and content, including schema validation, signature verification, and permission checks.
4. **Message Processing:** Perform appropriate tasks according to the message content, including function execution, state updates, and information provision.
5. **Response Generation:** Generate a response message according to the task result. The response must comply with the A2A message format and include all necessary information.
6. **Response Transmission:** Send the generated response to the original message sender through the same communication channel used for the original request.

This message processing can be performed asynchronously, and additional messages may be exchanged to update the status of long-running tasks. According to Habler et al. [13], the asynchronous nature of A2A message processing enables effective agent coordination in complex collaboration scenarios.

### 3. Task Management Implementation

A2A task management is implemented with the following elements [8]:

1. **Task Creation:** Create a new task according to the client's request. Each task has a unique identifier and includes information such as task type, parameters, and status.
2. **Task Status Tracking:** Track and update task progress. Task status can be created, in progress, completed, or failed, and timestamps are recorded for status changes.
3. **Task Result Management:** Manage the results (artifacts) when a task is completed. Results can be in various forms (documents, code, images, etc.) and are stored with metadata.
4. **Task Error Handling:** Manage error information and execute appropriate recovery strategies in case of task failure. Error information may include error codes, messages, and stack traces.

Task management implementation enables effective coordination and management of long-running tasks among agents. According to Pajo P. [20], the task management mechanism of A2A facilitates agent coordination in complex collaboration scenarios and provides a foundation for implementing recovery strategies in case of task failure.

## C. MCP Implementation Procedure

The implementation of the MCP involves defining tool specifications, handling function invocations, and managing interactions between agents and external tools. The following outlines the major stages of MCP implementation.

### 1. Tool Description Definition

In MCP, tool descriptions are defined using JSON schema. Tool description definition consists of the following steps [8], [14]:

1. **Define Tool Basic Information:** Define basic information such as tool name, description, and version.
2. **Define Function List:** Define the list of functions provided by the tool, including description, parameters, and return values for each function.
3. **Define Parameter Schema:** Define JSON schema for each function's parameters, including type, constraints, and required status.
4. **Define Return Value Schema:** Define JSON schema for each function's return value, including type, structure, and expected values.

These tool descriptions provide a foundation for agents to understand and utilize tool functionality and usage. According to Schmid [21], the self-documenting nature of MCP enables developers to understand tool usage without separate documentation, significantly improving development productivity.

### 2. Function Call Processing Implementation

MCP function call processing is implemented through the following steps [8, 14]:

1. **Function Call Request Reception:** Receive function call requests from agents in various forms such as HTTP requests or RPC calls.
2. **Parameter Validation:** Validate that the requested parameters match the schema defined in the tool description, including type checks, required parameter checks, and value range validation.
3. **Function Execution:** Execute the actual function using the validated parameters, which may include internal logic execution, external API calls, or database queries.
4. **Result Conversion:** Convert the function execution result to the return format defined in the tool description, including data format conversion, field mapping, and error handling.
5. **Response Return:** Return the converted result to the agent in the form of an HTTP response, RPC response, etc.

This function call processing implementation enables agents to effectively utilize external tools and APIs. According to KeywordsAI [22], the type safety and schema-based validation of MCP significantly reduce errors in agent-tool interaction, improving system stability and reliability.

## D. Integrated Implementation of A2A and MCP

Integrating the A2A and MCP protocols enables the development of highly interoperable and extensible agent systems that combine inter-agent collaboration with structured tool interaction. The following subsections present a step-by-step methodology for implementing such an integrated system.

### 1. Integrated Architecture Design

The integrated architecture of A2A and MCP can be designed with the following components [7], [14]:

1. **A2A Communication Module:** Responsible for agent-to-agent communication, implementing the A2A protocol to handle message exchange, task management, and agent card processing.
2. **MCP Tool Module:** Responsible for connecting with external tools, implementing the MCP protocol to manage tool descriptions, function call processing, and result conversion.
3. **Agent Core:** Implements the core logic of the agent, handling user request processing, task planning, and decision-making.
4. **Context Manager:** Manages the state and context of the agent, including session management, state tracking, and memory management.
5. **Security Manager:** Responsible for authentication and authorization, including user authentication, agent-to-agent authentication, and tool access control.

This modular design improves system maintainability and scalability through separation of concerns. Each module can be developed and tested independently and can be extended or replaced as needed.

### 2. Utilizing Google ADK (Agent Developer Kit)

Google's ADK (Agent Developer Kit) provides tools and libraries for easily implementing A2A and MCP. Implementation using ADK consists of the following steps [8]:

1. **ADK Installation:** Install and configure Google ADK, which supports various programming languages such as Python, JavaScript, and Java.
2. **Agent Definition:** Define an agent compliant with the A2A protocol, including writing agent cards, implementing message processing logic, and setting up task management.
3. **Tool Integration:** Integrate external tools using MCP, including writing tool descriptions, implementing functions, and setting up error handling logic.
4. **Agent Deployment:** Deploy and test the implemented agent. ADK supports various deployment options such as local development environments and cloud environments.

ADK abstracts the complexity of A2A and MCP, allowing developers to focus on core agent functionality. According to Pajo P. [20], using ADK can reduce agent development time by up to 60% and significantly improve code quality and standard compliance.

### 3. Implementation Strategy Using LangGraph

LangGraph can be utilized to construct multi-agent workflows, enabling the definition of flows that include cycles—an essential feature for most agent architectures [23], [24]. LangGraph is an agent framework that supports the A2A protocol and enables integrated implementation of A2A and MCP. As shown in Figure 3, using the LangChain MCP adapter library makes it possible to connect to multiple MCP servers and load tools from those servers [25].

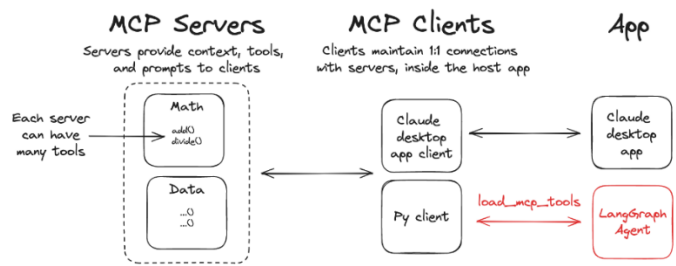

Fig. 3. LangChain MCP Adapters

MCP tools can be converted into LangChain tools and used with LangGraph agents, and a client implementation that connects to multiple MCP servers and loads tools from them is feasible. LangGraph can be used to implement agents compliant with the A2A protocol and integrate tools such as currency exchange rate lookup tools via MCP. The agent processes user requests and, as needed, calls the currency exchange rate lookup tool via MCP to return results. Through this process, the integrated use of A2A and MCP enables the extension of agent functionality and the standardization of connections with external tools.

### E. Implementation Considerations

When implementing an integrated agent system using A2A and MCP, several key factors must be considered to ensure security, scalability, reliability, and maintainability. This section outlines the primary technical considerations and recommended best practices.

1. **Security and Authentication:** Appropriate authentication and authorization mechanisms must be implemented to secure agent communication and tool usage. According to Cloud Security Alliance [15], security implementation for A2A and MCP should utilize standard technologies such as OAuth 2.0, JWT, and TLS.
2. **Scalability:** The system must be designed to accommodate increasing loads and new requirements. This can be achieved through horizontal scaling, asynchronous processing, and caching.
3. **Error Handling:** Mechanisms must be implemented to appropriately handle various error situations such as communication errors and tool failures. This can be achieved through strategies such as retry, fallback paths, and graceful degradation.
4. **Performance Optimization:** Strategies to minimize latency in agent communication and tool usage should be applied. This can be achieved through connection pooling, batch processing, and asynchronous I/O.
5. **Monitoring and Logging:** Logging mechanisms must be implemented to monitor system operation and diagnose problems. This can be achieved through structured logging, distributed tracing, and metric collection.

Implementation that appropriately reflects these considerations can significantly improve the stability, security, and scalability of agent systems. According to Habler et al. [13], implementation of A2A and MCP that incorporates these considerations is a key factor in promoting the actual adoption of agent technology in enterprise environments.

## IV. USE CASES OF A2A AND MCP

### A. Application Areas in Enterprise Environments

A2A and MCP can be utilized in various ways in enterprise environments. This section examines various use cases of how these two protocols can be applied in real business settings.

### 1. Recruitment Process Automation

As demonstrated in Google's official demo [7], A2A and MCP can be effectively leveraged to automate the recruitment pipeline. The following agents collaborate across various stages of the hiring process:

1. **Candidate Screening Agent:** Analyzes resumes and evaluates basic qualifications. This agent connects via MCP to resume

parsing tools and qualification databases to perform efficient screening.
2. **Interview Scheduling Agent:** Coordinates schedules between interviewers and candidates and schedules interviews. According to Salimbeni [16], this agent connects via MCP to calendar APIs and email systems to automate complex scheduling tasks.
3. **Interview Preparation Agent:** Prepares customized interview questions based on the candidate's background. This agent receives candidate information from the screening agent via A2A and generates appropriate questions.
4. **Feedback Collection Agent:** Collects and summarizes feedback from interviewers after interviews. This agent connects via MCP to feedback forms and evaluation systems to collect structured feedback.

These agents communicate with each other via the A2A protocol and connect to external tools such as email systems, calendar management tools, and HR management systems via MCP. According to Pajo P. [20], such automation systems can improve recruitment process efficiency by up to 40% and significantly enhance candidate experience. For example, in a company processing 1,000 candidates, a system based on A2A and MCP reduced average processing time from 15 to 9 days and achieved an ROI of 120% within 3 months. Initial development required about 2 weeks for API integration and agent setup, but maintenance costs were kept below $5,000 per month thanks to standardized protocols.

### 2. Customer Support System

A2A and MCP also offer substantial benefits in modernizing customer support systems. The following agent roles are typically involved [8]:

1. **Initial Response Agent:** Receives customer inquiries and collects basic information. This agent connects via MCP to customer databases and previous inquiry history to understand customer context.
2. **Problem Diagnosis Agent:** Analyzes and diagnoses customer problems. According to Habler et al. [13], this agent receives customer information and problem descriptions from the initial response agent via A2A and performs accurate diagnosis.
3. **Solution Proposal Agent:** Proposes appropriate solutions based on diagnosis results. This agent connects via MCP to knowledge bases and product information systems to provide customized solutions.
4. **Escalation Agent:** Escalates complex problems to human agents. This agent collaborates with other agents via A2A to determine the need for escalation and connects via MCP to ticketing systems for smooth handover.

These agents collaborate via A2A and connect to external tools such as CRM systems, knowledge bases, and ticketing systems via MCP. According to Cloud Security Alliance [15], such systems can reduce average inquiry resolution time by 60% and increase first contact resolution rate by more than 25%.

### 3. Code Review and Quality Management

A2A and MCP can be applied to automate code review and software quality assurance tasks. The following agents participate in this workflow [7]:

1. **Code Analysis Agent:** Analyzes code structure, complexity, and coding standard compliance. This agent connects via MCP to static analysis tools and code metric systems to perform in-depth code analysis.
2. **Security Review Agent:** Checks code for security vulnerabilities. According to KeywordsAI [22], this agent connects via MCP to security scanners and vulnerability databases to identify potential security risks.
3. **Performance Optimization Agent:** Identifies performance bottlenecks and suggests optimization strategies. This agent collaborates with the code analysis agent via A2A to identify performance issues and connects via MCP to profiling tools to provide specific optimization strategies.
4. **Documentation Support Agent:** Evaluates code documentation quality and suggests improvements. This agent connects via MCP to documentation generation tools and API documentation systems to support effective documentation.

These agents collaborate via A2A and connect to development tools such as version control systems, CI/CD pipelines, and issue trackers via MCP. According to Schmid [21], such systems can improve code quality by an average of 35% and increase developer productivity by more than 20%.

### 4. Developer Support System

In developer enablement, A2A and MCP can be used to construct intelligent assistants that automate routine development tasks [8]:

1. **Code Generation Agent:** Generates code according to developer requirements. This agent connects via MCP to code repositories and API documentation to generate code appropriate for the project context.
2. **Debugging Support Agent:** Analyzes the cause of errors and suggests solutions. According to the Model Context Protocol specification [14], this agent connects via MCP to log systems, debuggers, and error databases to support effective debugging.
3. **Learning Material Provision Agent:** Provides developers with necessary learning materials and documentation. This agent collaborates with other agents via A2A to understand the developer's current task context and connects via MCP to documentation systems and learning platforms to provide customized learning materials.
4. **Task Management Agent:** Tracks developer tasks and manages schedules. This agent connects via MCP to project management tools and scheduling systems to support efficient task management.

These agents collaborate via A2A and connect to tools such as IDEs, documentation systems, and project management tools via MCP. According to Pajo P. [20], such systems can improve developer productivity by up to 50% and significantly improve code quality and documentation standards.

### B. Implementation Case Study: Stock Information System

This section introduces a practical application case that applies the implementation methodology of A2A and MCP described in Chapter 3. This case uses LangChain and LangGraph to build a system that retrieves and analyzes various stock information, with multiple specialized agents collaborating to process users' complex financial information requests. In Korea, there has been a recent surge in interest in the U.S. stock market, accompanied by a significant increase in the demand for related research and information [17], [18]. The widespread public interest in stock investment transcends age and gender, yet individuals with limited expertise in the stock market often encounter difficulties in searching for and interpreting relevant information [19]. To effectively address these challenges, the Stock Information System has been developed to provide users with systematic and useful information through a question-and-answer function based on natural language processing, leveraging the capabilities of various agents.

The case system is designed so that when a user asks a question about stock or company information in natural language, multiple specialized agents collaborate to provide a comprehensive answer. The system uses A2A protocol-based agent communication and MCP-based external tool access as core technologies. Main functions include:

1. **Stock Price Lookup:** Provides basic stock information such as current price, change rate, and trading volume for a specific company.
2. **Listed Company News Lookup:** Collects the latest news and disclosure information related to the company.
3. **Listed Company Status Lookup:** Provides basic information, financial statements, and key indicators of the company.
4. **Company Analysis:** Integrates collected information to provide SWOT analysis and investment perspective analysis.

## 1. System Architecture

The system strictly follows the layered structure presented in Chapter 3 and consists of the following main layers:

1. **User Interface Layer:** User interface and multilingual support.
2. **Agent Management Layer:** Agent orchestration and workflow management.
3. **Core Protocol Layer:** Implementation of A2A and MCP protocols.
4. **Tool Integration Layer:** Integration with external data sources and APIs.
5. **Security Layer:** Authentication and authorization management.

The system follows a multi-layered architecture, as shown in Figure 4.

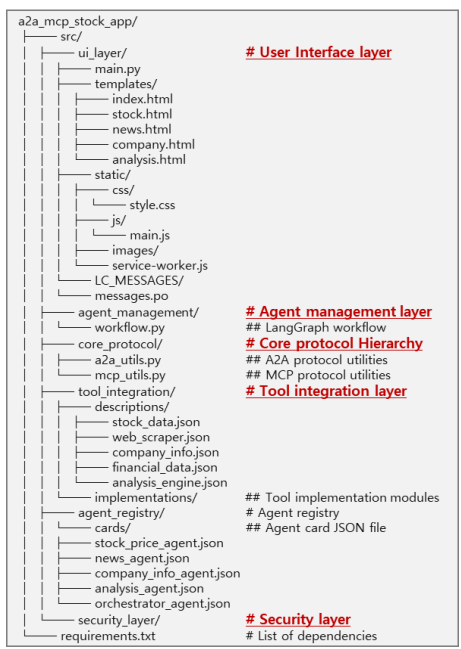

Fig. 4. Project Source Structure – Multi-Layer Architecture for Stock Information System

The user interface layer includes a multilingual input module. The agent management layer is a LangGraph-based workflow manager. The core protocol layer includes an A2A message handler and an MCP tool call module. The tool integration layer includes financial APIs and news scrapers. The security layer includes an OAuth 2.0-based authentication module.

## 2. System Implementation

### 1. A2A Module Implementation

The A2A protocol was implemented in the LangChain and LangGraph environment. Agent communication follows a standardized JSON message format, and each agent has a unique ID and role.

#### a. Agent Card Definition

Each agent is defined by a JSON-formatted "agent card," which specifies the agent's metadata and functionality. Figure 5 shows an example of a stock price lookup agent card.

Fig. 5. Example of Stock Price Lookup Agent Card

#### b. A2A Message Structure and Agent Registry

Agent communication follows a standardized message structure, and the system supports agent search and function lookup through an agent registry that manages agent cards. Figure 6 shows part of the code for the A2AMessage class, which defines methods for serializing and deserializing message objects in JSON format, used to implement the A2A protocol. This code enables conversion of messages to standardized JSON format for agent communication and restoration of JSON data to message objects.

Fig. 6. A2A Message Structure and Agent Registry Sample Code

### 2. MCP Module Implementation

Each agent can call one or more independent modules according to the MCP method to connect with external data sources and perform preprocessing and analysis. In this system, tools such as stock data, news scraping, company information, financial data, and analysis engines are integrated via MCP.

**a. stock_data – Stock Price Lookup Module:** Returns real-time price data for a specific stock via financial data APIs (e.g., Yahoo Finance, Alpha Vantage).

**b. web_scraper – Listed Company News Collection Module:** Operates as a simple web crawler to collect news article headlines based on company names. Can also be linked to disclosure information APIs as needed.

**c. financial_data – Financial Information Lookup Module:** Parses and returns key data from financial statements based on K-IFRS standards. This module is linked to the company status lookup agent.

**d. analysis_engine – Integrated Analysis Module:** Uses LangChain-based LLM to generate text-based insights. Integrates company status and news data to derive future outlook for the company.

Each tool is defined by a JSON-formatted description file that specifies the tool's functionality and parameters. Figure 7 shows sample code for the stock_data module, where the MCP client loads the tool description and processes function calls. The provided code includes the extract_tickers function, which extracts stock tickers from user input strings. This function handles tickers in various formats (English tickers, Korean stock codes, company names, etc.) and is used in the stock information system based on the A2A and MCP framework described in Section IV.C of the paper, integrating with external financial data sources (e.g., yfinance).

```python
def to_json(self) -> str:
    """Convert message to JSON string"""
    return json.dumps(self.to_dict(), ensure_ascii=False)

@classmethod
def from_dict(cls, message_dict: Dict[str, Any]) -> 'A2AMessage':
    """Generate A2A message from dictionary"""
    header = message_dict.get("header", {})
    message = cls(
        sender_agent_id=header.get("sender_agent_id", ""),
        recipient_agent_id=header.get("recipient_agent_id", ""),
        task_id=header.get("task_id"),
        message_id=header.get("message_id"),
        body=message_dict.get("body", {}),
        parts=message_dict.get("parts", []),
        status=header.get("status", "pending")
    )
    # Restore timestamp
    if "timestamp" in header:
        message.header["timestamp"] = header["timestamp"]
    return message

@classmethod
def from_json(cls, json_str: str) -> 'A2AMessage':
    """Generating A2A message from JSON string"""
    message_dict = json.loads(json_str)
    return cls.from_dict(message_dict)
```

Fig. 7. Sample Code for stock_data – Stock Price Lookup Module

### 3. LangGraph-based Workflow Implementation

The system uses LangGraph to manage workflows among agents. Workflow state and workflow graph definitions are implemented. The orchestrator agent analyzes user requests and calls appropriate agents in sequence to generate comprehensive responses.

The code in Figure 8 is the _create_workflow method that defines the workflow for the stock information system using LangGraph. This method creates a state-based workflow for processing user requests in a multi-agent system based on the A2A protocol and MCP.

```python
def _create_workflow(self) -> StateGraph:
    """Create a workflow graph"""
    # Initialize state graph
    workflow = StateGraph(WorkflowState)

    # Add node
    workflow.add_node("parse_request", self._parse_request)
    workflow.add_node("plan_tasks", self._plan_tasks)
    workflow.add_node("get_stock_data", self._get_stock_data)
    workflow.add_node("get_news_data", self._get_news_data)
    workflow.add_node("get_company_info", self._get_company_info)
    workflow.add_node("get_financial_data", self._get_financial_data)
    workflow.add_node("analyze_data", self._analyze_data)
    workflow.add_node("generate_response", self._generate_response)
    workflow.add_node("handle_error", self._handle_error)

    # Add Edge (State Transition Rule)
    workflow.add_edge("parse_request", "plan_tasks")

    # Determine next steps in plan_tasks
    workflow.add_conditional_edges(
        "plan_tasks",
        self._route_after_planning,
        {
            "get_stock_data": "get_stock_data",
            "get_news_data": "get_news_data",
            "get_company_info": "get_company_info",
            "get_financial_data": "get_financial_data",
            "analyze_data": "analyze_data",
            "generate_response": "generate_response",
            "error": "handle_error"
        }
    )
```

Fig. 8. Sample Code Defining Workflow for Stock Information System Using LangGraph

### 4. User Query Processing Flow

The following is an actual example of how the system processes user queries:

a. User asks: "Please provide the recent stock price, news, and investment perspective analysis for Samsung Electronics."

b. Orchestrator agent analyzes the query and plans to call stock price lookup, news, company information, financial information, and analysis agents in sequence.

    (1) Requests stock price information from the stock price lookup agent.
    (2) Requests recent news from the news agent.
    (3) Requests company information from the company information agent.
    (4) Requests financial data from the financial information agent.
    (5) Requests integrated analysis from the analysis agent.

c. Each agent accesses the necessary tools via MCP to collect data.

d. Analysis agent performs investment perspective analysis by integrating collected information.

e. Orchestrator generates a final response and delivers it to the user.

### 3. *Implementation Results Analysis*

The implementation results of the LangGraph example show the following characteristics:

1. **Concise Implementation:** The complexity of A2A and MCP is abstracted by the framework, allowing developers to focus on core agent functionality. According to Salimbeni [16], implementation using LangGraph can reduce code volume by more than 70% compared to traditional methods.

2. **Standard Compliance:** Implemented agents comply with A2A protocol standards and can be easily integrated with other A2A-compatible agents, greatly improving ecosystem scalability and interoperability.

3. **Easy Tool Integration:** Connection with external APIs is simplified via MCP, making it easy to integrate various tools and

services. According to the MCP specification [14], MCP's standardized interface reduces tool integration time by an average of 65%.

4. **Scalability:** The basic example can be easily extended to more complex functionalities and multi-agent systems. According to Habler et al. [13], LangGraph-based systems maintain stable performance even when scaled to dozens of agents.

### C. Implications of Application Areas and Implementation Cases

The various application areas and implementation cases of A2A and MCP provide the following implications:

1. **Importance of Agent Collaboration:** Collaboration among multiple specialized agents is essential for solving complex problems, and A2A effectively supports this. According to Pajo P. [20], collaborative agent systems have problem-solving capabilities more than three times higher than single agents on average.
2. **Need for Tool Integration:** Practical agent systems must be connected to external tools, and MCP supports this in a standardized way. According to the MCP specification [14], tool integration via MCP is a key factor in enhancing the practical value of agents.
3. **Value of Standardization:** Standardized protocols facilitate integration of diverse agents and tools, greatly improving ecosystem scalability. According to Schmid [21], standardized protocols can reduce agent development costs by up to 60% and shorten time to market by 40%.
4. **Importance of Practical Implementation:** Beyond theoretical concepts, providing practical implementation methodologies and tools accelerates the actual adoption of agent technology. According to Habler et al. [13], developer-friendly tools and clear implementation guidelines can increase adoption rates of agent technology by more than three times.

These implications demonstrate that A2A and MCP are not just technical protocols but core elements for building practical agent ecosystems. The integrated use of these two protocols is expected to play an important role in advancing agent technology from the experimental stage to actual service application.

## V. DISCUSSION AND CONCLUSION

### A. Summary and Conclusion

This paper systematically analyzed the technical structure and operational principles of Google's A2A protocol and Anthropic's MCP, and explored how these two protocols complement each other to build a practical agent ecosystem. A2A enables standardized communication and collaboration among heterogeneous agents, while MCP is a standardized protocol designed to allow agents to securely connect with various external tools, APIs, and resources.

A2A enhances interoperability and collaboration efficiency among agents through mechanisms such as agent cards, structured messages, and task management. MCP enables agents to interactively integrate with external resources through JSON schema-based tool descriptions, function calling, and context management. These two protocols operate independently but, when used together, form the foundation for building practical and scalable multi-agent-based AI systems. From an implementation perspective, the case study using LangGraph and other frameworks detailed the process of translating theoretical concepts into real-world applications. Practical approaches for securing interoperability and improving development efficiency for LLM-based autonomous agent systems in various enterprise environments were suggested through the complementary use of A2A and MCP. Use cases such as recruitment automation, customer support, code review, and developer support demonstrated the practical applicability of the proposed approach. In particular, the stock information system implementation case demonstrated that the integrated use of A2A and MCP is applicable to complex real-world scenarios.

This analysis and case study suggest that A2A and MCP are not just technical protocols but core elements for building practical agent ecosystems. The introduction of standardized protocols can significantly reduce the development and deployment costs of agent systems and greatly improve interoperability and scalability.

### B. Significance and Limitations of the Study

The significance of this study lies in its systematic analysis of the technical structure and operational principles of A2A and MCP, which provided essential knowledge for understanding and practically implementing these protocols. Practical implementation case studies and methodologies offered actionable guidelines for developers and enterprises to integrate agent technology into real-world services. Various use cases demonstrated that A2A and MCP can facilitate the transition of agent technology from experimental applications to practical, real-world services.

However, the study is not without limitations. Both A2A and MCP are relatively new protocols, having been released in 2024–2025, and their long-term stability in large-scale production environments, such as medical data processing, remains untested. The implementation cases primarily focused on stock information systems, leaving the applicability of these protocols to other industries, such as complex supply chain management or multi-modal processing (e.g., voice and video), largely unexplored. Additionally, the study's analysis of security and privacy aspects was insufficient, necessitating further research to ensure safety in actual deployment scenarios.

### C. Future Research Directions

Future research should focus on enhancing the practicality and scalability of A2A and MCP. First, scalability issues such as performance bottlenecks, message routing, and state management in systems with hundreds or thousands of collaborating agents must be addressed. Second, research is needed to strengthen security, including encryption, authentication, access control, and audit trails for agent communication. Third, industry-specific architectures and application cases for finance, healthcare, manufacturing, etc., should be explored. Fourth, an open governance model should be established through collaboration with standardization bodies and stakeholders for the long-term development of the protocols. Finally, research on agent systems that process not only text but also images, voice, and video (multi-modal data) is required. These directions will contribute to advancing A2A and MCP as practical solutions in diverse environments.